\documentclass[sigconf]{acmart}

\copyrightyear{2023}
\acmYear{2023}
\setcopyright{acmlicensed}
\acmConference[MM '23] {Proceedings of the 31st ACM International Conference on Multimedia}{October 29--November 3, 2023}{Ottawa, ON, Canada.}
\acmBooktitle{Proceedings of the 31st ACM International Conference on Multimedia (MM '23), October 29--November 3, 2023, Ottawa, ON, Canada}
\acmPrice{15.00}
\acmISBN{979-8-4007-0108-5/23/10}
\acmDOI{10.1145/3581783.3611833}

\usepackage{epstopdf}
\usepackage{graphicx}
\usepackage{amsmath}

\usepackage{booktabs}
\usepackage[ruled]{algorithm2e}
\usepackage{algpseudocode}
\usepackage{multirow}
\usepackage{bbding}

\usepackage{amssymb}

\DeclareMathOperator*{\argmin}{arg\,min}

\usepackage[capitalize]{cleveref}
\crefname{section}{Sec.}{Secs.}
\Crefname{section}{Section}{Sections}
\Crefname{table}{Table}{Tables}
\crefname{table}{Tab.}{Tabs.}
\crefname{algocf}{alg.}{algs.}

\begin{document}

\title{Unlearnable Examples Give a False Sense of Security:\\
Piercing through Unexploitable Data with Learnable Examples}

\author{Wan Jiang}
\authornote{Both authors contributed equally to this research.}
\affiliation{%
  \institution{Hefei University of Technology}
  \city{Hefei}
  \country{China}
}

\author{Yunfeng Diao}
\authornotemark[1]
\authornote{Corresponding authors. E-mails: diaoyunfeng@hfut.edu.cn, he{\_}wang@ucl.ac.uk}
\affiliation{%
  \institution{Hefei University of Technology}
  \city{Hefei}
  \country{China}
}

\author{He Wang}
\authornotemark[2]
\affiliation{%
  \institution{University College London}
  \city{London}
  \country{United Kingdom}}

\author{Jianxin Sun}
\affiliation{%
  \institution{Chinese Academy of Science}
  \city{Beijing}
  \country{China}
}

\author{Meng Wang}
\affiliation{%
 \institution{Hefei University of Technology}
 \city{Hefei}
 \country{China}}

\author{Richang Hong}
\affiliation{%
 \institution{Hefei University of Technology}
 \city{Hefei}
 \country{China}}

\renewcommand{\shortauthors}{Wan Jiang et al.}
\begin{abstract}

Safeguarding data from unauthorized exploitation is vital for privacy and security, especially in recent rampant research in security breach such as adversarial/membership attacks. To this end, \textit{unlearnable examples} (UEs) have been recently proposed as a compelling protection, by adding imperceptible perturbation to data so that models trained on them cannot classify them accurately on original clean distribution. Unfortunately, we find UEs provide a false sense of security, because they cannot stop unauthorized users from utilizing other unprotected data to remove the protection, by turning unlearnable data into learnable again. Motivated by this observation, we formally define a new threat by introducing \textit{learnable unauthorized examples} (LEs) which are UEs with their protection removed. The core of this approach is a novel purification process that projects UEs onto the manifold of LEs. This is realized by a new joint-conditional diffusion model which denoises UEs conditioned on the pixel and perceptual similarity between UEs and LEs. Extensive experiments demonstrate that LE delivers state-of-the-art countering performance against both supervised UEs and unsupervised UEs in various scenarios, which is the first generalizable countermeasure to UEs across supervised learning and unsupervised learning. Our code is available at \url{https://github.com/jiangw-0/LE_JCDP}.

\end{abstract}

\ccsdesc[500]{Security and privacy~Human and societal aspects of security and privacy}
\ccsdesc[500]{Computing methodologies~Machine learning}

\keywords{Unlearnable Examples, Data Protection, Deep Neural Network}


\maketitle

\section{Introduction}
The abundance of ``freely'' accessible data on the Web has been pivotal to the success of modern deep learning, such as ImageNet~\cite{russakovsky2015imagenet} and Ms-celeb-1m~\cite{Guo2016MSCeleb1MAD}. However, these datasets might include personal data collected without mutual consent~\cite{birhane2021large}, which has raised public concerns that private data can be utilized to create commercial models without the owner's authorization~\cite{hill2019photos}. To address such concerns, growing efforts~\cite{RUE,HuangUE,yu2022availability} have been made to add protection to data to prevent unauthorized usage by making the data unexploitable. These methods add imperceptible ``shortcut'' noise to the images so that the deep learning models learn no useful semantics but correspondences between noise and labels~\cite{geirhos2020shortcut}. Consequently, the models trained on unexploitable data fail to classify clean data, thereby safeguarding users' privacy. Such poisoning methods are named as unlearnable example (UE) protection~\cite{HuangUE} or availability attack~\cite{yu2022availability}. 


While the growing research focuses on how to make data unexploitable~\cite{sankar2023cuda,zhang2022unlearnable,ren2022transferable,fowl2021adversarial,wang2021fooling,he2022indiscriminate}, we aim to challenge this paradigm by exposing a key vulnerability in this protection: the protection is merely effective if the unexploitable data is all that is accessible. Unfortunately, this is often not the case. Data protectors can only add the ``unlearnable” perturbations to their own data, but they cannot prevent unauthorized users from accessing similar, unprotected data from other sources. As a result, one can study the underlying distribution of the protected examples, via studying similar newly collected (unprotected) data. Taking face recognition as an example, although unlearnable examples cannot be directly used to train classifiers, it is easy to collect new unprotected face data. As long as there is sufficient similarity between the newly collected (unprotected) data and the original clean data, it is still possible to train a classifier that can successfully classify the original clean data. In other words, unauthorized users can easily bypass data protection to learn the original data representation from newly collected unprotected data, even if the data might be small in scale, different from the clean data, lacks label annotation, and is alone not ideal for training a classifier~\cite{wang2018transferring,closer_fewshot}.

To show the existence of the aforementioned vulnerability, we design a new approach that can turn unlearnable examples into learnable ones. A straightforward solution would be to design a specific training scheme that can train on unexploitable data~\cite{MadryAT,qin2023learning,liu2023image}. This is less ideal as it merely classifies unexploitable data but not reveal much about the underlying clean data, i.e. unprotected version of the unlearnable data. We argue that an ultimate countermeasure is to infer/expose the underlying clean data by turning UEs into learnable again, which can enable further unauthorized exploitation such as standard training or representation learning~\cite{ren2022transferable,he2022indiscriminate}. Therefore, the \textit{learnable unauthorized data} should be independent of training scheme and can be normally used just like original training data. We refer to examples in \textit{learnable unauthorized data} as \textit{learnable examples} (LEs). The key idea behind obtaining \textit{learnable examples} is to learn a learnable data manifold from other similar data and then project unlearnable examples onto this manifold.

Inspired by the power of the diffusion models in noise purification~\cite{nie2022diffusion} and image generation~\cite{dhariwal2021diffusion}, we propose a novel purification method based on diffusion models, called joint-conditional diffusion purification, to capture the mapping from the unlearnable examples to their corresponding clean samples. We first inject the unlearnable images with controlled amounts of Gaussian noises progressively, until their unlearnable perturbations are submerged by Gaussian noise. Next, we equip the denoising process with a new joint condition that speeds up noise removal while preserving image semantics. The joint condition is parameterized by both the pixel distance and the neural perception distance between the unlearnable sample and its corresponding denoised version. This is based on the observation that unlearnable examples typically exhibit small differences in pixel distance from clean samples and the differences are imperceptible to the human vision. Therefore, the denoised images should closely resemble the original samples through minimizing the visual difference from the unlearnable example. 


We extensively evaluate our approach on both supervised and unsupervised UEs across a number of benchmark datasets, and compare it with existing countering methods. The results show LE substantially outperforms existing countermeasures and it is the only one that maintains effectiveness under both supervised learning and unsupervised learning. More importantly, unlike existing countermeasures that are tied to specific training schemes, our learnable examples are independent of them and can be used normally as the original clean training data. Surprisingly, we found that our approach still retains effectiveness even when there is a large distributional difference between the newly collected data (utilized in training a learnable data manifold) and the clean data. In other words, the distributions between training data and collected raw data can be different and we can still turn unlearnable examples into learnable. This undoubtedly further deepens our concerns about the vulnerability of unexploitable data since it does not require the collected raw data to be very similar to the unprotected version of the unlearnable examples.

In summary, our main contributions are: 1) We identify and demonstrate an inherent vulnerability of UE protection, by formally defining an ultimate threat to UEs called \textit{learnable examples}, which can turn unlearnable examples into learnable ones. 2) We propose a novel purification strategy for producing \textit{learnable examples}, called Joint-conditional Diffusion Purification, which purifies UEs with a diffusion model simultaneously conditioned on pixel and perceptual similarity. 3) We demonstrate that LE outperforms existing state-of-the-art countermeasures against both supervised UEs and unsupervised UEs. LE is the first generalizable countermeasure across supervised learning and unsupervised learning. 4) We empirically demonstrate that the joint-conditional diffusion model can still purify UEs even when the learned density is not the same as the clean distribution, exposing the fragility of protection by UEs.


     
    


\section{Related Work}
\subsection{Unlearnable Examples}
Unlearnable examples (UEs) are a type of poisoning methods~\cite{poisoning}, but aimed at defending against unauthorized data exploitation~\cite{HuangUE,munoz2017towards}. The vanilla UEs define a bi-level optimization objective, which makes the optimal solution on UEs have a maximum loss on clean data~\cite{HuangUE,deepconfuse,ntga}. Given its effectiveness and efficiency, many variants of UEs have been proposed, such as robust UEs~\cite{RUE,wen2023adversarial}, manually designed UEs~\cite{sankar2023cuda,yu2022availability}, clustering-based UEs~\cite{zhang2022unlearnable}, and sparse UEs~\cite{onepixel}. Inspired by the effectiveness of UEs in supervised learning, He et al.~\cite{advcl} and Ren et al.~\cite{ren2022transferable} investigate the impact of UEs on unsupervised learning.  

Countermeasures against UEs have only been attempted very recently~\cite{qin2023learning,liu2023image, dolatabadi2023devil}. Adversarial Training (AT)~\cite{HuangUE} has been shown to partially resist UE protection, but robust UE soon broke through this countermeasure~\cite{RUE,wen2023adversarial}. Adversarial augmentation~\cite{qin2023learning}, which combines various data augmentation policies with adversarial training, is further proposed to improve generalization of the unauthorized model. Liu et al.~\cite{liu2023image} suggest using grayscale transformation to counter UE protection. However, these methods are associated with specific training schemes, which limits the use of unauthorized data for other training schemes and tasks. The recent arXiv paper~\cite{dolatabadi2023devil} applies diffusion models to counter UEs. The major differences to our approach are that we propose a new joint-conditional diffusion model instead of a naive application of the diffusion model, tackling the trade-off between perturbation purification and image semantic retaining. Furthermore, all prior countermeasures~\cite{HuangUE,qin2023learning,liu2023image, dolatabadi2023devil} are designed for supervised learning, but not much is known about the fragility of unsupervised UEs.

\subsection{Diffusion Models}
Diffusion models~\cite{ddpm,sohl2015deep} have surpassed Generative Adversarial Networks (GANs)~\cite{gans} in the field of image generation and have achieved impressive results with conditions such as text, semantic maps or reference images~\cite{latent-ddpm,avrahami2022blended,diffusion-auto}. A typical diffusion model consists of two processes: a forward process and a denoising process. The former gradually approaches Gaussian noise by iteratively adding noise to clean images, while the latter obtains real images from noise in the form of Markov chains. Since a carefully designed denoising process can defuse the ramifications of data perturbations, recent works~\cite{nie2022diffusion,wang2022guided,may2023salient,dolatabadi2023devil} leverage the power of diffusion models for noise purification, such as adversarial purification. However, the key difference between existing noise purification and our work is the accessibility of training data. Existing purification methods~\cite{nie2022diffusion,dolatabadi2023devil} assume that training data (the unprotected version of the unlearnable examples) is available for training, allowing diffusion models to learn the original data distribution in advance. However, UE makes the training data unexploitable. How to train a diffusion model(and more generally learning a similar data manifold) without access to training data poses a crucial challenge for removing UE protection. Existing purification methods have yet to explore this aspect.



\begin{figure*}[ht]
\centering
\includegraphics[width=0.8\linewidth]{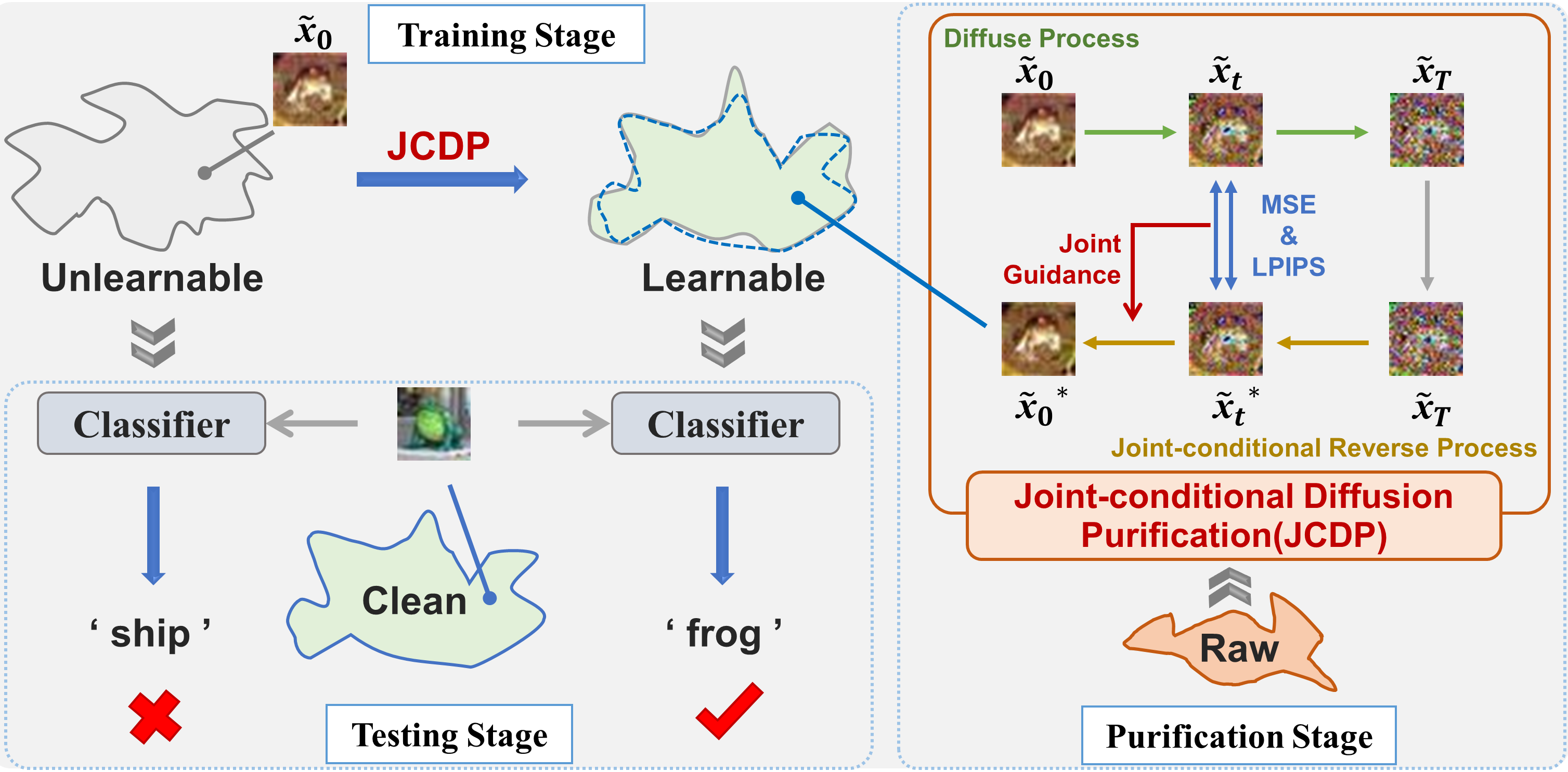}
\caption{An illustration of producing \textit{learnable examples} via joint-conditional diffusion purification (JCDP). The classifier training on unlearnable data does not generalize well on clean distribution. To remove such data protection, we propose JCDP that projects unlearnable data onto a learnable data manifold, which is learned from newly collected raw (unprotected) data.}
\label{fig:overview}

\end{figure*} 
\section{Methodology}
\subsection{Problem Statement}
In this subsection, we first give the definition of UEs, and then define the typical countermeasures against UE protection.  
\subsubsection{Data Protector.} Suppose that the data protectors have access to the original dataset and it is denoted as $\mathcal{D}_{c}=\{(\boldsymbol{x}^{(i)},y^{(i)})\}^{N}_{i=1}$ with $N$ clean samples, in which i.i.d. input-label pairs $(\boldsymbol{x}^{(i)},y^{(i)})$ are drawn from the joint data distribution $p_{d}(\boldsymbol{x},y)$. The protectors' goal is to protect the data from unauthorized training after its release. To this end, they release an unlearnable dataset $\mathcal{D}_{u}=\{(\Tilde{\boldsymbol{x}}^{(i)},y^{(i)})\}^{N}_{i=1}$ to the unauthorized users. The protector aims to make a classifier $f_{\theta}:\mathcal{X} \rightarrow \mathcal{Y}$ trained on the unlearnable dataset perform poorly on the original clean data distribution $p_{d}(\boldsymbol{x},y)$:
\begin{align}\label{eq:UE}
    \boldsymbol{\theta}^{*} = \argmin_{\boldsymbol{\theta}}\mathbb{E}_{(\Tilde{\boldsymbol{x}}, y) \in \mathcal{D}_{u}}[\mathcal{L}(f_{\boldsymbol{\theta}}(\Tilde{\boldsymbol{x}}), y)] \\\nonumber
    \text{s.t.}~p_{\boldsymbol{\theta}^{*}}(y|\boldsymbol{x}) \neq p_{d}(y|\boldsymbol{x})
\end{align}
where $\mathcal{L}(\cdot)$ is the cross-entropy loss. Since unlearnable perturbations should not affect the normal data utility, it is assumed that $\tilde{\boldsymbol{x}} = \boldsymbol{x} + \boldsymbol{\delta}$, where 
$\boldsymbol{\delta}$ is the ``invisible'' unlearnable perturbations bounded by $||\boldsymbol{\delta}||_{p} \leq \varepsilon$.

\subsubsection{Unauthorized Data Exploiter.} We assume the unauthorized parties only have access to the protected data, i.e. unlearnable examples set $\mathcal{D}_{u}$. Their goal is to train models on $\mathcal{D}_{u}$ and make them generalize well on the original clean data distribution. To this end, existing countermeasures~\cite{HuangUE,qin2023learning,liu2023image} have attempted to design a special training scheme that can train classifiers on unexploitable data. It has been shown that adversarial training (AT)~\cite{MadryAT} can be used to prevent UE protection to some extent~\cite{HuangUE}, which is formulated as follows:
\begin{equation}
    \argmin_{\boldsymbol{\theta}}\mathbb{E}_{(\Tilde{\boldsymbol{x}}, y) \in \mathcal{D}_{u}}\Big[ \max_{||\boldsymbol{\delta_{adv}}||_{p} \leq \varepsilon} \mathcal{L}(f_{\boldsymbol{\theta}}(\Tilde{\boldsymbol{x}}+\boldsymbol{\delta}_{adv}), y)\Big]
    \label{eq:at}
\end{equation}

Inspired by AT, adversarial augmentations (AA)~\cite{qin2023learning} is further proposed very recently. Specially, they combine data augmentation policies with adversarial training:
\begin{equation}
    \argmin_{\boldsymbol{\theta}}\mathbb{E}_{(\Tilde{\boldsymbol{x}}, y) \in \mathcal{D}_{u}}\Big[ \max_{\mathcal{T}\sim \mathcal{A}} \mathcal{L}(f_{\boldsymbol{\theta}}(\mathcal{T}(\Tilde{\boldsymbol{x}}), y)\Big]
    \label{eq:aa}
\end{equation}
where $\mathcal{T}(\cdot)$ is the combination of image augmentation policies from a set of all possible data augmentations $\mathcal{A}$. However, AT-based methods suffer from a significant performance drop when training on robust unexploitable data~\cite{RUE}. Moreover, AT modifies the training schemes, leaving the unauthorized data itself still unexploitable. This forces the training data to be tied to adversarial training, which limits the use of data for other training schemes and tasks, such as standard training and representation learning. Last but not least, using AT to train a large model on large-scale unexploitable datasets is not desirable, given the computational complexity involved~\cite{wongfast}.


\subsection{Learnable Examples}
Existing countering methods rely on specific training schemes. However, we argue that an ultimate threat against UEs should turn unexploitable data $\mathcal{D}_{u}$ into a learnable dataset $\mathcal{D}_{l}$. $\mathcal{D}_{l}$ can be used normally as the original clean training data, enabling further unauthorized exploitation such as standard straining and representation learning. Here we consider standard setting in supervised learning as an example, a model trained on $\mathcal{D}_{l}$ can easily generalize well on the original clean data distribution $p_{d}(\boldsymbol{x},y)$:
\begin{align}\label{eq:LE}
    \boldsymbol{\theta}^{*} = \argmin_{\boldsymbol{\theta}}\mathbb{E}_{(\boldsymbol{x}, y) \in \mathcal{D}_{l}}[\mathcal{L}(f_{\boldsymbol{\theta}}(\boldsymbol{x}), y)] \\\nonumber
    \text{s.t.}~\mathcal{D}_{l} = \texttt{denoise}(\mathcal{D}_{u}); \, p_{\boldsymbol{\theta}^{*}}(y|\boldsymbol{x}) = p_{d}(y|\boldsymbol{x}) 
\end{align}

We refer to learnable protected examples in $\mathcal{D}_{l}$ as \textit{learnable examples} (LEs). Note that LEs are independent of training schemes, hence they also can be used for unauthorized unsupervised learning. Given a good denoiser $\texttt{denoise}$, we can project unlearnable examples back to the learnable data manifold to obtain corresponding LEs. A good denoiser is often achieved by learning a generative model $G$~\cite{nie2022diffusion}. However, how to train a generator without access to the original training data is a tricky challenge. Our key observation is that although the original training data $\mathcal{D}_{c}$ is not accessible, small-scale raw (unprotected) data without labeled annotation $\mathcal{D}_{r}$ can be easily collected in the wild~\cite{henzler2021unsupervised}. As long as there is sufficient similarity between $\mathcal{D}_{r}$ and $\mathcal{D}_{u}$, we can learn an unconditional generator $G$ from $\mathcal{D}_{r}$ and utilize it to project UEs onto the manifold of LEs. Considering that diffusion models~\cite{ddpm,sohl2015deep} can achieve a high sample quality~\cite{dhariwal2021diffusion,sde} and noise purifying performance~\cite{nie2022diffusion}, we use a diffusion model for denoising. However, a naive application of the diffusion model will suffer the trade-off between noise purification and image semantic retaining. To tackle this problem, we propose a new joint-conditional diffusion purification conditioned on simultaneously measuring the pixel and perceptual similarity between UEs and corresponding denoised ones. An overview illustration is shown in \cref{fig:overview}. Next, we give details of the purification process.

\subsection{Joint-conditional Diffusion Purification}
\subsubsection{DDPM for Data Purified.}
Diffusion model defines a Markov chain of diffusion steps to add Gaussian noise gradually to the data and then learn the reversal of the diffusion process to construct desired data samples from the noise. Given a clean data point $\boldsymbol{x}_{0}$ sampled from a data distribution $q(\boldsymbol{x}_{0})$, Denoising Diffusion Probabilistic Models (DDPM)~\cite{ddpm} define a forward or diffusion process that follows Markov chain to gradually add Gaussian Noise to $\boldsymbol{x}_{0}$ in $T$ steps with pre-defined variance schedule $\beta_{1:T} \in (0,1)^T$:
\begin{align}
\label{eq:forward}
    q(\boldsymbol{x}_{1:T}|\boldsymbol{x}_{0}) = \prod^{T}_{t=1} q(\boldsymbol{x}_{t}|\boldsymbol{x}_{t-1})\\ \nonumber
    q(\boldsymbol{x}_t|\boldsymbol{x}_{t-1}) = \mathcal{N}(\boldsymbol{x}_{t};\sqrt{1-\beta_{t}}\boldsymbol{x}_{t-1},\beta_{t}\boldsymbol{I})
\end{align}
If we define $\alpha_{t}=1-\beta_{t},\overline{\alpha}_{t}=\prod^{t}_{s=1}\alpha_{s}$, we can reformulate the diffuse process via a single step:
\begin{align}
\label{eq:qt}
    q(\boldsymbol{x}_{t}|\boldsymbol{x}_{0}) = \mathcal{N}(\boldsymbol{x}_{t};\sqrt{\overline{\alpha}_{t}} \boldsymbol{x}_{0},(1-\overline{\alpha}_{t})\boldsymbol{I})
\end{align}
The reverse process is also a Markov process that learning a model $p_{\varphi}$ to estimate these conditional probabilities $q(\boldsymbol{x}_{t-1}|\boldsymbol{x}_{t})$:
\begin{align}
\label{eq:reverse}
    p_{\varphi}(\boldsymbol{x}_{0:T}) = p(\boldsymbol{x}_{T})\prod^{T}_{t=1} p_{\varphi}(\mathbf{x}_{t-1}|\boldsymbol{x}_{t})\\ \nonumber
    p_{\varphi}(\boldsymbol{x}_{t-1}|\boldsymbol{x}_{t}) = \mathcal{N}(\boldsymbol{x}_{t-1};\boldsymbol{\mu}_{\varphi}(\boldsymbol{x}_t,t),\boldsymbol{\sigma}_t^2\boldsymbol I)
\end{align}
where the reverse process is started from $p(\boldsymbol{x}_T)=\mathcal{N}(\boldsymbol{x}_T;\boldsymbol{0},\boldsymbol{I})$. The mean $\boldsymbol{\mu}_{\varphi}(\boldsymbol{x}_t,t)$ is a neural network parameterized by $\varphi$, while the variance $\boldsymbol{\sigma}_t^2$ can be learned by a neural network~\cite{improvedddpm} or a set of time-dependent constants~\cite{ddpm}. 

Now we introduce the purification process via utilizing DDPM. Assume that we have trained a DDPM model on collected raw data $\mathcal{D}_{r}$. Suppose the unlearnable example $\Tilde{\boldsymbol{x}}_{0}= \boldsymbol{x}_0 + \boldsymbol{\delta}$, we first diffuse the unlearnable image for $T_p$ steps by adding Gaussian noise to submerge the unlearnable perturbation:
\begin{equation}
\label{eq:xt}
    \Tilde{\boldsymbol{x}}_{T_p} = \sqrt{\overline{\alpha}_{T_p}} \boldsymbol{x}_0 +\sqrt{\overline{\alpha}_{T_p}} \boldsymbol{\delta} + \sqrt{1-\overline{\alpha}_{T_p}}\boldsymbol{\epsilon}
\end{equation}
where $\boldsymbol{\epsilon}$ is a standard Gaussian noise. The reverse process aims to simultaneously mitigate the added Gaussian noise and the residual unlearnable perturbation in $\Tilde{\boldsymbol{x}}_{T_p}$:
\begin{equation}
\label{eq:xt-1}
    \Tilde{\boldsymbol{x}}_{t-1} ^* =\frac{1}{\sqrt{\alpha_t}}(\Tilde{\boldsymbol{x}}_t ^*  - \frac{1-\alpha_{t}}{\sqrt{1-\overline{\alpha_t}}}\boldsymbol{\epsilon}_{\varphi}(\Tilde{\boldsymbol{x}}_t ^* ,t)) + \boldsymbol{\sigma}_t \boldsymbol \epsilon_t
\end{equation}
where $\Tilde{\boldsymbol{x}}_{T_p} ^* = \Tilde{\boldsymbol{x}}_{T_p}$. During the purification process described above, selecting the optimal purification step $T_p$ is crucial. If $T_p$ is too small, the unlearnable perturbation term $\sqrt{\overline{\alpha}_{T_p}} \boldsymbol{\delta}$ cannot be fully submerged by Gaussian noise, while choosing too large $T_p$ will lead to a loss of original semantic information. In our preliminary experiments, we have found that iteratively purifying the unlearnable images multiple times can mitigate this problem to some extent. Choosing a relatively small $T_p$ to conduct multiple purification iterations is more effective than purifying once with a large $T_p$. The detailed iteration process is listed in \cref{alg1}.



\subsubsection{Joint-conditional Diffusion Model.}
To further eliminate the trade-off between purification strength and semantic content retaining, we propose a novel purification process based on diffusion model, called Joint-conditional Diffusion Purification (JCDP). JCDP leverages both pixel and perception distance guidance to enable joint control on low-level and high-level semantic similarity between the purified image to the original clean one. From the perspective of the unauthorized parties, the original clean training data is not known a \textit{priori}, we hence turn to retain consistency between the purified image and the unlearnable one during the reverse process. This approximation is reasonable because unlearnable example by definition is imperceptible to humans, and its unlearnable perturbation is constrained to a small $\epsilon$-ball by pixel distance. Therefore, the purified image will closely resemble its original clean sample by encouraging it to be visually close to the unlearnable example.

The pixel distance can be easily calculated by mean square error (MSE). However, the true perception distance cannot be directly computed for image data. Considering that perceptual similarity can be intuitively linked to deep visual representation~\cite{LPIPS}, we propose to use neural perception distance LPIPS~\cite{LPIPS} to approximate the true perception distance. We introduce an extra neural network $h(\cdot)$ and denote its generated feature embeddings as $\Phi(\boldsymbol{x})$. Next, according to condition this reverse process $p_{\varphi}(\boldsymbol{x}_{t-1}|\boldsymbol{x}_{t})$ in \cref{eq:reverse} on $\Tilde{\boldsymbol{x}}_t $ and $\Phi({\Tilde{\boldsymbol{x}}_t})$, we can obtain the denoised learnable sample from the joint conditional distribution $p_{\varphi}(\Tilde{\boldsymbol{x}}_{0:T_p}^*|\Tilde{\boldsymbol{x}}, \Phi({\Tilde{\boldsymbol{x}}}))$: 
\begin{align}
\setlength{\abovedisplayskip}{3pt}
\setlength{\belowdisplayskip}{3pt}
\label{eq:reverse_joint}
    p_{\varphi}(\Tilde{\boldsymbol{x}}_{0:T_p}^*|\Tilde{\boldsymbol{x}}, \Phi({\Tilde{\boldsymbol{x}}})) = p(\Tilde{\boldsymbol{x}}_{T_p}^*)\prod^{T}_{t=1} p_{\varphi}(\Tilde{\boldsymbol{x}}_{t-1}^*|\Tilde{\boldsymbol{x}}_{t}^*, \Tilde{\boldsymbol{x}}_t, \Phi({\Tilde{\boldsymbol{x}}_t}))
\end{align}
where $\Tilde{\boldsymbol{x}}_t$ is the noise version of $\Tilde{\boldsymbol{x}}_0$ via $t$-step diffusion. Using Bayes' rule, we can derive the joint conditional reverse process:
\begin{equation}
\begin{split}
\label{eq:joint}
    &p_{\varphi}(\Tilde{\boldsymbol{x}}_{t-1}^* |\Tilde{\boldsymbol{x}}_{t}^* , \Tilde{\boldsymbol{x}}_t, \Phi({\Tilde{\boldsymbol{x}}_t})) \\
    & =  \frac{p(\Phi({\Tilde{\boldsymbol{x}}_t})|\Tilde{\boldsymbol{x}}_{t-1}^* , \Tilde{\boldsymbol{x}}_t^* , \Tilde{\boldsymbol{x}}_t ) p_{\varphi}(\Tilde{\boldsymbol{x}}_{t-1}^* |\boldsymbol{x}_{t}^* , \Tilde{\boldsymbol{x}}_t )}{p( \Phi({\Tilde{\boldsymbol{x}}_t})| \Tilde{\boldsymbol{x}}_t^* , \Tilde{\boldsymbol{x}}_t )} \\
    & = p(  \Tilde{\boldsymbol{x}}_{t}^*    |\Tilde{\boldsymbol{x}}_{t-1}^* ,  \Tilde{\boldsymbol{x}}_t, \Phi({\Tilde{\boldsymbol{x}}_t}) )\frac{p(\Phi({\Tilde{\boldsymbol{x}}_t})|\Tilde{\boldsymbol{x}}_{t-1}^* , \Tilde{\boldsymbol{x}}_t )}{p( \Tilde{\boldsymbol{x}}_{t}^* |\Tilde{\boldsymbol{x}}_{t-1}^*, \Tilde{\boldsymbol{x}}_t )} \frac{p_{\varphi} (\Tilde{\boldsymbol{x}}_{t-1}^* |\Tilde{\boldsymbol{x}}_{t}^*, \Tilde{\boldsymbol{x}}_t )}{p( \Phi({\Tilde{\boldsymbol{x}}_t})|\Tilde{\boldsymbol{x}}_{t}^*, \Tilde{\boldsymbol{x}}_t )}
\end{split}
\end{equation}
Since the diffuse process of DDPM is a Markov process, we have: 
\begin{equation}
\begin{split}
\label{eq:jointequ}
    p_{\varphi}(\Tilde{\boldsymbol{x}}_{t-1}^*|\Tilde{\boldsymbol{x}}_{t}^*, \Tilde{\boldsymbol{x}}_t, \Phi({\Tilde{\boldsymbol{x}}_t})) \!= \!\frac{p(\Phi({\Tilde{\boldsymbol{x}}_t})|\Tilde{\boldsymbol{x}}_{t-1}^*, \Tilde{\boldsymbol{x}}_t ) p_{\varphi}(\Tilde{\boldsymbol{x}}_{t-1}^* | \Tilde{\boldsymbol{x}}_{t}^*, \Tilde{\boldsymbol{x}}_t )} {p(\Phi({\Tilde{\boldsymbol{x}}_t})|\Tilde{\boldsymbol{x}}_{t}^*, \Tilde{\boldsymbol{x}}_t )}
\end{split}
\end{equation}
then we take the logarithm of both sides 
\begin{equation}
\begin{split}
\label{eq:logjoint}
    {\rm log}~ p_{\varphi}(\Tilde{\boldsymbol{x}}_{t-1}^* & |\Tilde{\boldsymbol{x}}_{t}^*, ~\Tilde{\boldsymbol{x}}_t, \Phi({\Tilde{\boldsymbol{x}}_t})) = {\rm log}~ p(\Phi({\Tilde{\boldsymbol{x}}_t})|\Tilde{\boldsymbol{x}}_{t-1}^* , \Tilde{\boldsymbol{x}}_t ) \\
    &+ {\rm log}~ p_{\varphi}(\Tilde{\boldsymbol{x}}_{t-1}^*| \Tilde{\boldsymbol{x}}_{t}^*, \Tilde{\boldsymbol{x}}_t )- {\rm log}~ p( \Phi({\Tilde{\boldsymbol{x}}_t})|\Tilde{\boldsymbol{x}}_{t}^*, \Tilde{\boldsymbol{x}}_t)\\
    \end{split}
\end{equation}
We can approximate ${\rm log}~ p( \Phi({\Tilde{\boldsymbol{x}}_t})|\Tilde{\boldsymbol{x}}_{t}^*, \Tilde{\boldsymbol{x}}_t)$ using a Taylor expansion around $\Tilde{\boldsymbol{x}}_{t}^* =\Tilde{\boldsymbol{x}}_{t-1}^*$. This gives
\begin{equation}
\begin{split}
\label{eq:logTaylor}
     &{\rm log}~ p_{\varphi}(\Tilde{\boldsymbol{x}}_{t-1}^*|\Tilde{\boldsymbol{x}}_{t}^*, \Tilde{\boldsymbol{x}}_t, \Phi({\Tilde{\boldsymbol{x}}_t})) \\
     &\! \thickapprox \! (\Tilde{\boldsymbol{x}}_{t-1}^* - \Tilde{\boldsymbol{x}}_{t}^* ) \nabla_{ \Tilde{\boldsymbol{x}}_{t}^*} {\rm log}~p(\Phi({\Tilde{\boldsymbol{x}}_t})|\Tilde{\boldsymbol{x}}_{t}^* , \Tilde{\boldsymbol{x}}_t)+ {\rm log}~ p_{\varphi}(\Tilde{\boldsymbol{x}}_{t-1}^* | \Tilde{\boldsymbol{x}}_{t}^*, \Tilde{\boldsymbol{x}}_t )
    \end{split}
\end{equation}
According to work \cite{sohl2015deep} and  \cite{dhariwal2021diffusion}, it is proved that:
\begin{equation}
\begin{split}
    &p_{\varphi}(\Tilde{\boldsymbol{x}}_{t-1}^*  | \Tilde{\boldsymbol{x}}_{t}^*, \Tilde{\boldsymbol{x}}_t ) \\
    &\thickapprox \mathcal{N}(\Tilde{\boldsymbol{x}}_{t-1}^* ;\boldsymbol{\mu}_{\varphi}(\Tilde{\boldsymbol{x}}_{t}^*, t) + \boldsymbol{\sigma}_t^2 \nabla_{\Tilde{\boldsymbol{x}}_{t}^*}\log p(\Tilde{\boldsymbol{x}}_t|\Tilde{\boldsymbol{x}}_{t}^* ), \boldsymbol{\sigma}_t^2 \boldsymbol I)  
\end{split}
\end{equation}
then we can show that 
\begin{equation}
\begin{split}
\label{eq:jointn}
    &p_{\varphi}(\Tilde{\boldsymbol{x}}_{t-1}^*|\Tilde{\boldsymbol{x}}_{t}^*, \Tilde{\boldsymbol{x}}_t, \Phi({\Tilde{\boldsymbol{x}}_t}))\\
    &\thickapprox \mathcal{N}(\Tilde{\boldsymbol{x}}_{t-1}^*;\boldsymbol{\mu}_{\varphi} (\Tilde{\boldsymbol{x}}_{t}^*, t) + \boldsymbol{\sigma}_t^2 (\boldsymbol{d}_1 + \boldsymbol{d}_2 ) ,\boldsymbol{\sigma}_t^2\boldsymbol I)\\ 
    &\boldsymbol{d}_1 = \nabla_{\Tilde{\boldsymbol{x}}_{t}^*} \log p(\Tilde{\boldsymbol{x}}_t|\Tilde{\boldsymbol{x}}_{t}^*), \boldsymbol{d}_2 =\nabla_{\Tilde{\boldsymbol{x}}_{t}^*} \log p(\Phi({\Tilde{\boldsymbol{x}}_t})| \Tilde{\boldsymbol{x}}_{t}^*, \Tilde{\boldsymbol{x}}_t )
    \end{split}
\end{equation}
In the above equation, $p(\Tilde{\boldsymbol{x}}_t|\Tilde{\boldsymbol{x}}_{t}^*)$ and $p( \Phi({\Tilde{\boldsymbol{x}}_t})|\Tilde{\boldsymbol{x}}_{t}^*, \Tilde{\boldsymbol{x}}_t)$ represent how likely $\Tilde{\boldsymbol{x}}_{t}^*$ is close to $\Tilde{\boldsymbol{x}}_t$ under the data space and latent feature space in the reverse process. Following \cite{wang2022guided}, we can utilize MSE as the distance metric $\boldsymbol {\mathcal{D}_m}$ to approximate $p(\Tilde{\boldsymbol{x}}_t|\Tilde{\boldsymbol{x}}_{t}^*)$:
\begin{align}
\label{eq:d1z}
    p(\Tilde{\boldsymbol{x}}_t|\Tilde{\boldsymbol{x}}_{t}^*) = \frac{1}{\boldsymbol Z}exp( \lambda_1 \boldsymbol {\mathcal{D}_m}(\Tilde{\boldsymbol{x}}_{t}^*, \Tilde{\boldsymbol{x}}_t))
\end{align}
where $\boldsymbol Z$ is a normalization factor and $\lambda_1$ is a scale factor that modulates the guidance strength. From \cref{eq:d1z} we have 

\begin{equation}
\boldsymbol{d}_1 = \nabla_{\Tilde{\boldsymbol{x}}_{t}^*} \log p(\Tilde{\boldsymbol{x}}_t|\Tilde{\boldsymbol{x}}_{t}^*) = -\lambda_1 \nabla_{\Tilde{\boldsymbol{x}}_{t}^*}\boldsymbol {\mathcal{D}_m}(\Tilde{\boldsymbol{x}}_{t}^*, \Tilde{\boldsymbol{x}}_t)
\end{equation}
Similarly, we quantify $p( \Phi({\Tilde{\boldsymbol{x}}_t})|\boldsymbol{x}_{t})$ by adapting LPIPS as the perception distance metric $\boldsymbol {\mathcal{D}_p}$:
\begin{align}
\label{eq:d2z}
    p(\Phi({\Tilde{\boldsymbol{x}}_t})| \Tilde{\boldsymbol{x}}_{t}^*, \Tilde{\boldsymbol{x}}_t ) = \frac{1}{\boldsymbol Z}exp( \lambda_2 \boldsymbol {\mathcal{D}_p}(\Tilde{\boldsymbol{x}}_{t}^*, {\Tilde{\boldsymbol{x}}_t})) \\ \nonumber
    \boldsymbol {\mathcal{D}_p}(\Tilde{\boldsymbol{x}}_{t}^*, {\Tilde{\boldsymbol{x}}_t}) = ||\Phi(\Tilde{\boldsymbol{x}}_{t}^*) - \Phi({\Tilde{\boldsymbol{x}}_t}) ||_2
\end{align}
Next, we can calculate its gradient
\begin{equation}
\boldsymbol{d}_2 = \nabla_{\Tilde{\boldsymbol{x}}_{t}^*} \log p(\Phi({\Tilde{\boldsymbol{x}}_t})| \Tilde{\boldsymbol{x}}_{t}^* , \Tilde{\boldsymbol{x}}_t) = -\lambda_2 \nabla_{\Tilde{\boldsymbol{x}}_{t}^*}\boldsymbol {\mathcal{D}_p}(\Tilde{\boldsymbol{x}}_{t}^*, {\Tilde{\boldsymbol{x}}_t})
\end{equation}

Based on the inference above, the conditional transition operator $p_{\varphi}(\Tilde{\boldsymbol{x}}_{t-1}^*|\Tilde{\boldsymbol{x}}_{t}^*, \Tilde{\boldsymbol{x}}_t, \Phi({\Tilde{\boldsymbol{x}}_t}))$ can be approximated by a Gaussian distribution, whose mean is shifted by $\boldsymbol{\sigma}_t^2 (\boldsymbol{d}_1 + \boldsymbol{d}_2)$. \cref{alg1} outlines the inference details of JCDP. 
\begin{algorithm}
\caption{Joint-conditional Diffusion Purification}
\label{alg1}
\LinesNumbered 
\KwIn{Unlearnable example $\Tilde{\boldsymbol{x}}_{0}$, diffusion step $T_p$ per each purification run, number of purification iterations $N$, given a DDPM $(\boldsymbol{\mu}_{\varphi}(\boldsymbol{x}_{t}, t), \boldsymbol{\sigma}_t^2\boldsymbol{I}) $, gradient scale $\lambda_1 $ and $\lambda_2$.}
\textbf{Init}: $\boldsymbol{d_1} = 0 $ and $\boldsymbol{d_2} = 0 $\;
\For {$i$ $\leftarrow 1$ to $N$}{
    The diffusion process: \\
    $q(\Tilde{\boldsymbol{x}}_{1:T_p}|\Tilde{\boldsymbol{x}}_{0}) = \prod^{T_p}_{t=1} q(\Tilde{\boldsymbol{x}}_{t}|\Tilde{\boldsymbol{x}}_{t-1})$;\\
    The reverse process: \\
    \For{$t \leftarrow T_p$ to $1$}{
    $\boldsymbol{\mu}, \boldsymbol{\sigma}_t^2  \leftarrow \boldsymbol{\mu}_{\varphi}(\Tilde{\boldsymbol{x}}_{t}^*, t), \boldsymbol{\sigma}_t^2 $\;
    $\boldsymbol{d_1} \leftarrow  -\lambda_1 \nabla_{\Tilde{\boldsymbol{x}}_{t}^*}\boldsymbol {\mathcal{D}_m}(\Tilde{\boldsymbol{x}}_{t}^*, \Tilde{\boldsymbol{x}}_t)$\;
    $\boldsymbol{d_2} \leftarrow  -\lambda_2 \nabla_{\Tilde{\boldsymbol{x}}_{t}^*} \boldsymbol {\mathcal{D}_p}(\Tilde{\boldsymbol{x}}_{t}^*, {\Tilde{\boldsymbol{x}}_t})$\;
    $\Tilde{\boldsymbol{x}}_{t-1}^* \leftarrow$ sample from $\mathcal{N} ( \boldsymbol{\mu} + \boldsymbol{\sigma}_t^2 (\boldsymbol{d}_1 + \boldsymbol{d}_2 ), \boldsymbol{\sigma}_t^2\boldsymbol{I})$
    }
}
\Return $\Tilde{\boldsymbol{x}}_{0}^*$
\end{algorithm}
\subsubsection{Fine-tuning for Diffusion Model.} We can directly train a diffusion model on newly collected (unprotected) data. However, since the size of the collected data is typically much smaller than the original training data, the generation quality may be low. Fortunately, it is almost impossible that data protectors add unlearnable perturbations to all data in the real world. Therefore, we can transfer knowledge from unprotected source domains to unauthorized target domains with limited collected data by means of fine-tuning. Specially, we first initialize DDPM with the weights of a source network pre-trained on one unprotected domain, and then fine-tune it on the newly collected data. This simple fine-tuning operation can shorten the convergence time and improve the purification performance, especially when collected raw data is limited.     

\section{Experiments}
\begin{table*}[!tb]
	\caption{Comparing test accuracy (\%) of Resnet-18 classifiers trained on protected datasets CIFAR-10, CIFAR-100 and SVHN.  Clean means standard training on clean training set. EMN refers to the sample-wise form of error-minimizing noise while EMN (S) is the class-wise type. Note that ISS contains 3 image transformation strategies and we report the best strategy. The best results are highlighted in bold.}
	\label{tab:main}
	\begin{center}		
		\resizebox{1\linewidth}{!}{
			\begin{tabular}{lcccccccccccccc}
				\toprule
				\multirow{2}{*}{\rotatebox[origin=c]{0}{{\textbf{Countermeasures}}}}
				&\multicolumn{5}{c}{\textbf{CIFAR-10~(Clean~95.3)}}
				&\multicolumn{5}{c}{\textbf{CIFAR-100~(Clean~78.8)}}
				&\multicolumn{4}{c}{\textbf{SVHN~(Clean~96.2)}}\\
				\cmidrule(lr){2-6} \cmidrule(lr){7-11} \cmidrule(lr){12-15}
				& EMN~\cite{HuangUE} & EMN (C)~\cite{HuangUE} & REMN~\cite{RUE}  & LSP~\cite{yu2022availability} &  TAP~\cite{fowl2021adversarial}& 
    
                EMN~\cite{HuangUE}  & EMN (C)~\cite{HuangUE}  &REMN~\cite{RUE}  & LSP~\cite{yu2022availability} & TAP~\cite{fowl2021adversarial}& 
                
                EMN~\cite{HuangUE} & REMN~\cite{RUE}  & LSP~\cite{yu2022availability}&
                TAP~\cite{fowl2021adversarial}\\
				\midrule
				Vanilla & 21.2 & 20.7 & 20.5 & 15.0 &7.8 & 
                    14.8 & 4.0 &10.9 & 4.1 & 8.6 & 
                    13.9 & 11.6 & 7.3 &41.0\\
                    
				AVATAR~\cite{dolatabadi2023devil} & 91.0 & - & 88.5 & 85.7& 90.7&
                65.7 & - & 64.9 & 58.5 &65.0 &
                93.8 & 88.5 & 83.8 &93.4 \\
                
				ISS~\cite{liu2023image}  &   93.0 & - &  $\textbf{92.8}$  & 82.5 &83.9 &
                67.5 & - & 57.3 & 53.5 & 56.3&
                89.9 & - & 92.2 &- \\
    
				AT~\cite{MadryAT}  &  84.8 & 85.0 & 49.2 & 80.2 &83.0&
                63.4 & 60.1 & 27.1 & 58.1 & 61.4&
                86.3 & 70.0 & 80.2 &-\\
    
				AA~\cite{qin2023learning}&   90.8 & - & 85.5 & 84.9 &- &
                70.0 &- & - & 67.4 & -&
                88.7 & - & 92.6&-\\
    
				LE (Ours) & $\textbf{93.1}$ &   $\textbf{94.0}$ &  92.2 &  $\textbf{92.4}$ & $\textbf{90.8}$ &
                $\textbf{70.9}$ &  $\textbf{67.8}$ &  $\textbf{65.3}$ &  $\textbf{68.7}$ &$\textbf{68.0}$ &
                $\textbf{94.7}$ &  $\textbf{89.9}$ &  $\textbf{93.3}$ &$\textbf{94.9}$\\
				\bottomrule
		\end{tabular}}
	\end{center}
\end{table*}
\subsection{Settings}
\label{sec:setting}
\subsubsection{Datasets and Evaluation.} We select three widely adopted benchmark datasets for LE evaluation: CIFAR-10~\cite{krizhevsky2009learning}, CIFAR-100~\cite{krizhevsky2009learning}, SVHN~\cite{netzer2011reading} and Pets~\cite{parkhi2012cats}. CIFAR-10\&100 consist of 50000 images in the training set and 10000 images in the test set. SVHN contains 73257 digit images for training and 26032 digit images for testing. Pets consists 3680 pet images for training and 3669 pet images for testing. To demonstrate the superiority of LE, we compare LE with AT~\cite{MadryAT} and the recently proposed supervised countermeasures, including ISS~\cite{liu2023image}, AA~\cite{qin2023learning} and AVATAR~\cite{dolatabadi2023devil}. AT, ISS and AA modify the training scheme. AVATAR uses pre-trained generative models to purify UEs and the used pre-trained models access the original training set. AA uses adversarial augmentation technology, which applies $K$ different random augmentations for each image. For a fair comparison, we do not consider adding extra data in training stage so we set $K=1$. 

To demonstrate LE is a generalizable countering approach under supervised and unsupervised learning, we employ state-of-the-art supervised UE and unsupervised UE for generating unexploitable data. The supervised UE protections include Error-Minimization Noise (EMN)~\cite{HuangUE}, Robust Error-Minimization Noise (REMN)~\cite{RUE}, Linear-separable Synthetic Perturbations (LSP)~\cite{yu2022availability} and Target Adversarial poisoning(TAP)~\cite{fowl2021adversarial}. For unsupervised UE, we employ Contrastive Poisoning (CP)~\cite{he2022indiscriminate} and Unlearnable Clusters (UC)~\cite{zhang2022unlearnable}. Following their default setting, we generated EMN, REMN and TAP on the backbone ResNet-18~\cite{he2016deep}, and CP on three unsupervised backbones, including SimCLR~\cite{simclr}, MoCo (v2)~\cite{moco} and BYOL~\cite{byol}. LSP is a manually designed UE so no backbone is required. UC is generated on backbone ResNet-18 base on surrogate model  and evaluated via an self-supervised SimCLR. Following the evaluation protocol in recent countermeasures~\cite{qin2023learning,liu2023image}, the perturbation radius is $l_{\infty}=8/255$ for EMN, REMN and TAP, $l_{2}=1.0$ for LSP and $l_{\infty}=16/255$ for UC. To achieve the strongest poison performance, all unlearnable approaches have a 100\% poisoning rate.
\subsubsection{Training Details.}
in the joint conditional diffusion purification process JCDP is a learning-free purification method, hence we need to train an unconditional DDPM from other unprotected data in advance. There are two settings for the collection of unprotected data: distribution match and distribution mismatch. In distribution match, we use the unprotected raw testing set (without label annotation), which belongs to the same distribution with the unprotected version of the unlearnable examples but without overlapping. Unless specified otherwise, we use it by default. The detailed settings of the distribution mismatch are reported in \cref{sec:dist}. ResNet-18~\cite{he2016deep} is used as the default classifier to train on purified data. Please refer to \textbf{supplementary document} for the fuller training settings and implemental details.

\subsection{Evaluation on Supervised UEs}

We report the comparison results with the state-of-the-art countermeasures~\cite{MadryAT,liu2023image,qin2023learning, dolatabadi2023devil} on all datasets in \cref{tab:main}. First, our proposed method achieves the best test accuracy in all scenarios (countermeasures vs. datasets vs. UE methods). The only exception is REMN on CIFAR-10, where there is only a 0.6\% difference with ISS. Secondly, it is apparent that other countering approaches exhibit the generalization problem across different UE methods. Specifically, AT and AA lead to a significant degradation in accuracy when countering REMN and LSP protection on CIFAR-10. LSP and TAP  also proves to be more resilient against AVATAR and ISS than other UE methods. However, our method, LE, does not have these issues, consistently performing well across all UE protection, demonstrating its effectiveness against unforeseen UE protection. This is because our approach aims to learn a good data representation that is independent of both UE and specific training schemes. Overall, existing countermeasures might not pose an ultimate threat to unexploitable data compared to LE. Next, we provide a detailed analysis. 
\paragraph{Comparsion with AT Methods and Data Augmentation.} 
AT~\cite{MadryAT} can be employed to counter UE protection. A recent study has shown that Adversarial Augmentation (AA)~\cite{qin2023learning}, which combines data augmentation with adversarial training, can further enhance the countering performance. However, this countermeasure heavily relies on the adversarial training procedure. The robust form of error-minimizing noise~\cite{RUE} can easily compromise their performance by replacing the normally trained surrogate in EMN with an adversarially trained model. As shown in~\cref{tab:main}, REMN can easily break AT and significantly degrade the performance of AA on CIFAR-10. In contrast, LE transfers well across different UE protections, whose performance on REMN is only slightly lower than EMN. Furthermore, the computational complexity of AT is typically higher than standard training. For LE, the unauthorized data only needs to be denoised once, then it can be used for training models in standard setting. Hence LE reduces training time by approximately 70\% compared with AT-based methods. We report the consuming time of producing learnable data using LE in supplementary document. In addition, considering that data augmentation technology is another commonly used countermeasure~\cite{dolatabadi2023devil,liu2023image}, we compare our approach with 4 common used data augmentations in \cref{tab:augment}. As shown in \cref{tab:augment}, LE significantly outperforms various data augmentations by a big margin. 
\begin{table}[!tb]
\caption{Comparsion with data augmentations on CIFAR10.}
\label{tab:augment}
\begin{center}
\begin{sc}
\resizebox{1\linewidth}{!}{
\setlength{\tabcolsep}{1 mm}{

\begin{tabular}{l|ccccccc}
\toprule
UE&Vanila &Cutout~\cite{cutout} &Mixup~\cite{mixup} &CutMix~\cite{cutmix} &FAutoAug.~\cite{fastaug} &LE (Ours)\\
\midrule
EMN &21.2  &23.8 &51.5 &25.3 &56.3 &\textbf{93.1}\\
REMN &20.5 &20.5 &26.6 &26.8 &26.6 &\textbf{92.2}\\
LSP &15.0 &11.8 &19.7 &23.8 &25.9 &\textbf{92.4}\\
\bottomrule
\end{tabular}
}}
\end{sc}
\end{center}
\end{table}

\paragraph{Comparsion with Image Compression.} Image Shortcut Squeezing (ISS) contains three common image compression operations, including grayscale transformation (GRAY), JPEG and bit depth reduction (BDR). We compare LE with every operation in ISS and the results are shown in~\cref{tab:iss}. Different from LS showing consistent countering performance across all UE protections, the optimal compression operation in ISS varies across different types of UEs, e.g. JPEG is not effective against error-minimizing noise (EMN, REMN) while GRAY is not effective against patch-based linear separable perturbation (LSP). In order to get the optimal performance, ISS has to ensemble multiple models which are applied with different compressions. As a result, LE is more efficient than ISS and has the best overall performance.

\begin{table}[!tb]
\caption{Comparsion with ISS on CIFAR-10 and CIFAR-100.}
\label{tab:iss}
\begin{center}
\begin{small}
\begin{sc}
\resizebox{1\linewidth}{!}{
\begin{tabular}{l|l|ccccc}
\toprule
Data&UE &Vanila &BGR&Gray&JPEG&LE (Ours)\\
\midrule
\multirow{4}{*}{CIFAR-10} &Clean  &94.7 &88.7 &92.4 &85.4 &n/a\\
& EMN &21.2 &36.5 &93.0 &81.5 &\textbf{93.1}\\
&REMN &20.5 &40.8 &\textbf{92.8} &82.3 &92.2\\
&LSP &15.0 &66.2 &82.5 &83.1 &\textbf{92.4}\\
\midrule
\multirow{4}{*}{CIFAR-100} &Clean &74.5 &-&71.8 &57.8 &n/a\\
& EMN &14.8 &- &67.5 &56.0 &\textbf{70.9}\\
&REMN &10.9 &-  &57.3 &55.8 &\textbf{65.3}\\
&LSP &4.1 &- &44.6 &53.5 &\textbf{68.7}\\
\bottomrule
\end{tabular}
}
\end{sc}
\end{small}
\end{center}
\end{table}

\paragraph{Comparsion with Noise Purification.} Similar to LE, AVATAR~\cite{dolatabadi2023devil} also uses diffusion model for purifying unlearnable perturbation. However, the main experiments reported in~\cite{dolatabadi2023devil} utilize the original clean training data for training diffusion models. In contrast, our approach does not need to have access to the unprotected version of the unlearnable examples (original training data), which is more available in real-world scenarios. In practice, the amount of newly collected (unprotected) data we used for training diffusion model is 10000 on CIFAR-10\&CIFAR-100, and 26032 on SVHN, much smaller than the amount of data used by AVATAR (50000 clean training data on CIFAR-10\&100 and 73257 on SVHN). In addition, unlike a naive application of diffusion model in AVATAR, we propose a novel joint-conditional diffusion method to improve the purification performance. The results in \cref{tab:main} show our proposed method surpasses AVATAR under all UE scenarios, especially in countering against LSP. It is worth noting that the amount of data used by LE is only 1/5 of that used by AVATAR on CIFAR-10\&100, and 1/3 on SVHN. We attribute the improvement to the new proposed joint-conditional diffusion model.  
\subsection{Evaluation on Unsupervised UEs}
\begin{table}[!tb]
    \caption{Performance of LE against unsupervised UE (CP\&UC). Table report linear probing accuracy. The best performance is shown in bold.}
	\label{tab:unsupervised}
	\begin{center}	
         \begin{small}
                \resizebox{1\linewidth}{!}{
			\begin{tabular}{cccccc}
				\toprule   
                    \multirow{1}{*}{\rotatebox[origin=c]{0}{{\textbf{UE}}}}&
				\multirow{1}{*}{\rotatebox[origin=c]{0}{{\textbf{Data}}}}
                &\multirow{1}{*}{\textbf{Backbone}}
				&\multirow{1}{*}{\textbf{Clean}}
				&\multirow{1}{*}{\textbf{Vanila}}&
                    \multirow{1}{*}{\textbf{LE (Ours)}}\\
                    \midrule
                    \multirow{6}{*}{\textbf{CP~\cite{he2022indiscriminate}}}
				&\multirow{3}{*}{\rotatebox[origin=c]{0}{CIFAR-10}}
				&SimCLR &90.4 &44.9 &\textbf{86.6}\\
                    &&MoCo v2 &89.3 &55.1 &\textbf{86.0}\\
                    &&BYOL &92.2 &59.6 &\textbf{85.7}\\
                    \cmidrule(lr){2-6}
                    &&SimCLR &63.6 &34.7 &\textbf{57.4}\\
                    &CIFAR-100&MoCo v2  &65.2 &41.9 &\textbf{57.1}\\
                    &&BYOL & 65.3 &39.2&\textbf{57.2}\\
                    \midrule
                    \multirow{1}{*}{\textbf{UC~\cite{zhang2022unlearnable}}}& \multirow{2}{*}{\rotatebox[origin=c]{0}{Pets}}& \multirow{2}{*}{SimCLR} & \multirow{2}{*}{48.3} &17.9 &  \textbf{46.0}  \\
                    
                    \multirow{1}{*}{\textbf{UC-CLIP~\cite{zhang2022unlearnable}}}& &  & &26.1 & \textbf{47.5} \\  
			    \bottomrule
			\end{tabular}}
        \end{small}
	\end{center}
\end{table}
Unlike existing countering methods, which are only available in supervised learning, LE is capable of countering unsupervised UEs as well. Considering the unsupervised UEs Contrastive Poisoning (CP)~\cite{he2022indiscriminate} are designed for unsupervised contrastive learning (UCL), we evaluate the effectiveness of LE on three well-known UCL frameworks, including SimCLR~\cite{simclr}, MoCo (v2)~\cite{moco} and BYOL~\cite{byol}. Unlearnable Cluster (UC)~\cite{zhang2022unlearnable} methods are not specifically designed for unsupervised setting, but are robustness to unsupervised exploitation against SimCLR. We evaluate the effectiveness of LE on  SimCLR for UC. As shown in~\cref{tab:unsupervised}, LE is consistently effective across various UCL algorithms. As there is no unsupervised countering method for direct comparison, we employ the contrastive learning version of AT (AdvCL)~\cite{advcl} and common data augmentations like Cutout~\cite{cutout} and Random Noise. We conduct comparisons using backbone SimCLR on CIFAR-10. The linear probing accuracy of AdvCL, Cutout and Random Noise is 79.3\%, 47.7\%, and 54.1\% respectively, while LE achieves 86.6\%, showing the superiority of LE by big margins in the context of unsupervised learning. To the best of our knowledge, LE is the first generalizable countermeasure that is effective against UEs in both supervised and unsupervised learning.
   
\subsection{Investigating the Distribution Similarity} 
\label{sec:dist}
Although LE requires collecting raw unprotected data to learn a data manifold, in practice, we find collecting raw data is not difficult because the distributions of the newly collected data(surrogate distribution) can be different from the original clean distribution. To evaluate LE's tolerance to distribution mismatch, we propose to estimate the scale of distribution mismatch using semantic similarity. We set up 3 scenarios with varying levels of semantic similarities, including: (1) high semantic similarity, (2) medium semantic similarity and (3) low semantic similarity. In scenario (1), we train DDPM on clean CIFAR-100 data and use it to purify the unlearnable CIFAR-10 training set. This is because a class with high similarity to CIFAR-10 classes can always be found in CIFAR-100. For example, although there is no ``dog'' (CIFAR-10) in CIFAR-100 class set, the CIFAR-100 class set contains other semantically similar animal classes such as ``fox'' and ``raccoon''. In scenario (2), we train DDPM on clean CIFAR-10 data and use it to purify the unlearnable CIFAR-100 dataset. For example, the classes in CIFAR-10 have a relatively large semantic difference from ``people'' classes in CIFAR-100 (``baby'', ``boy'', ``girl'', ``man'', and ``woman''). We define such semantic similarity as medium level. In scenario (3), we train DDPM on clean SVHN and purify on unlearnable CIFAR-10\&100. Scenario (3) is the most extreme case since the door digits in SVHN are totally different from the objects in CIFAR-10\&100. 

\begin{table}[!tb]
\caption{The results using different surrogate distributions. S-Distribution means the surrogate distribution for training DDPM.}
\label{tab:mismatch}
\begin{center}
\begin{small}
\begin{sc}
\begin{tabular}{lcccc}
\toprule
Setting&Data&S-Distribution &EMN&LSP\\
\midrule
\multirow{2}{*}{(1)} &\multirow{2}{*}{CIFAR-10}  &Vanilla &21.2 &15.0\\
&&CIFAR-100 &92.1 &89.1\\
\midrule
\multirow{2}{*}{(2)} &\multirow{2}{*}{CIFAR-100}  &Vanilla &14.8 &4.1\\
&&CIFAR-10 &66.9 &66.0\\
\midrule
\multirow{2}{*}{(3)} &CIFAR-10  &SVHN &89.3 &85.6\\
&CIFAR-100  &SVHN &52.9 &54.1\\
\bottomrule
\end{tabular}
\end{sc}
\end{small}
\end{center}
\end{table}

\begin{figure}[!tb]
\centering
\includegraphics[width=\linewidth]{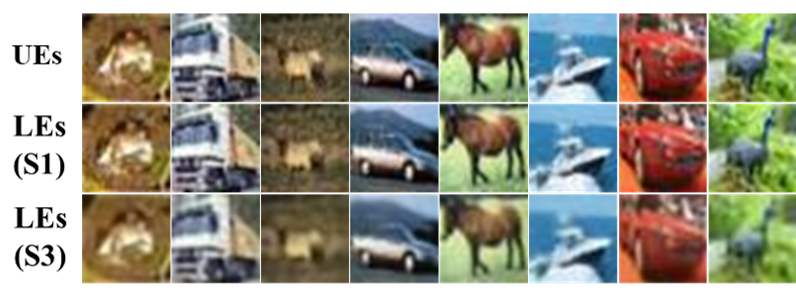}
\caption{Visual results of LEs with distribution mismatch. UEs are drawn from the protected (unlearnable) CIFAR-10 dataset. LEs(S1) and LEs(S3) choose the surrogate distribution CIFAR-100 and SVHN respectively.}
\label{fig:dist}
\end{figure}

For a fair comparison, we do not use fine-tuning and keep the amount of data for training DDPM consistent with the setting in~\cref{sec:setting}. The results are presented in \cref{tab:mismatch}. When the learned density is not far from the original clean distribution(scenario 1\&2), LE can still achieve high test accuracy. Surprisingly, even when the learned data manifold is totally different from the original data distribution (scenario 3), LE can still largely improve the test accuracy. To further understand why LEs can tolerate distribution mismatch, we visualize some examples under scenario 1 and 3 in~\cref{fig:dist}. Even under extreme scenario 3, LEs(S3) still retain the main original semantic features, albeit more blurred than corresponding UEs and LEs(S1). We speculate that this is because the joint-conditional terms in the reverse process (\cref{eq:jointn}) shift the mean of the learned data distribution (SVHN), pulling the purified image towards the CIFAR-10 distribution. It is hard to theoretically identify the cause and we will leave it to future research. Overall, this discovery further shatters the illusion of protected data, as it is impractical to add unlearnable noise to all unprotected images in the real world.

\subsection{Ablation Studies}
\begin{table}[!tb]                                        
\caption{Ablation analysis LE on CIFAR-10 and CIFAR-100. ``FT'' means whether to use fine-tuning for training DDPM, and ``JC'' means whether to use Joint-conditional Diffusion Purification. STEPS means the training steps of DDPM(batch size 256)}
\label{tab:ablation}
\begin{center}
\begin{small}
\begin{sc}
\begin{tabular}{lll|cc}
\toprule
FT&JC &Steps&EMN(CIFAR-10) &EMN(CIFAR-100)\\
\midrule
\XSolidBrush  &\XSolidBrush &80000 &90.6 &69.0\\
\Checkmark &\XSolidBrush &1000 &91.4 &69.3\\
\Checkmark &\XSolidBrush &10000 &91.9 &69.7\\
\Checkmark &\Checkmark &10000 &\textbf{93.1} &\textbf{70.9}\\
\bottomrule
\end{tabular}
\end{sc}
\end{small}
\end{center}
\end{table}

\begin{figure}[!tb]
\centering
\includegraphics[width=\linewidth]{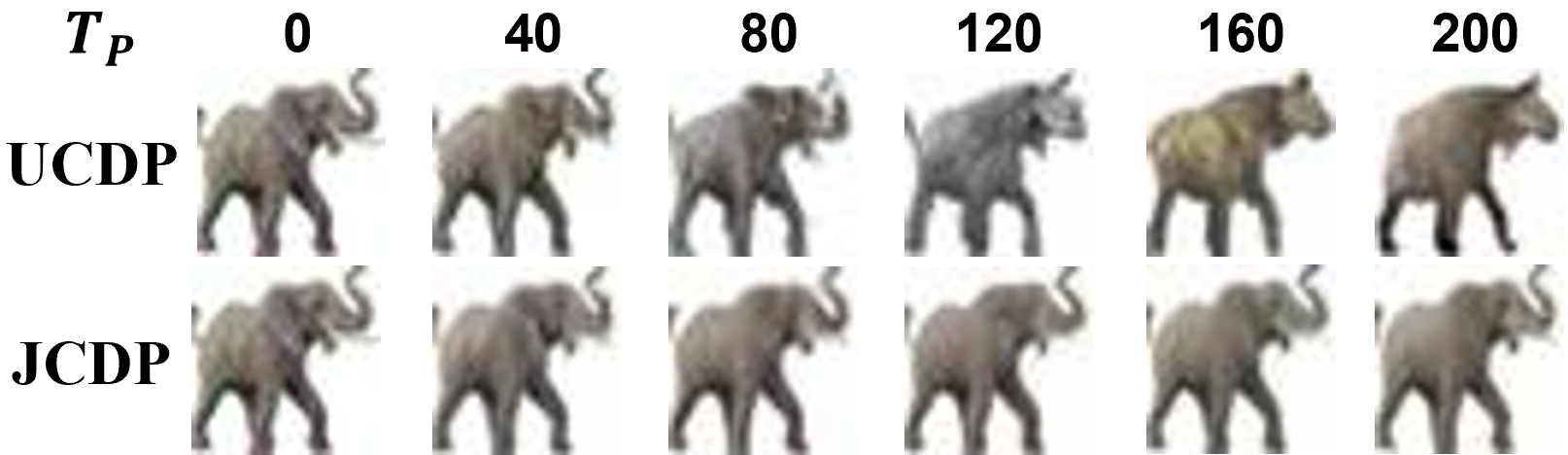}
\caption{Visual examples of Joint-conditional Diffusion Purification(JCDP) vs. Unconditional Diffusion Purification(UCDP) on CIFAR-100.}
\label{fig:abl}
\end{figure}

To further understand the comparative effects of different elements of our proposed method, we conduct two pairs of ablation studies: Fine-tuning vs. Training from Scratch, and Joint-conditional Diffusion Purification vs. Unconditional Diffusion Purification. ``Training from Scratch'' means directly training DDPM on collected raw data, while fine-tuning means fine-tuning from a DDPM trained on other unprotected datasets. The results are shown in~\cref{tab:ablation}. Compared with training from scratch, fine-tuning DDPM can generate higher-quality images already in earlier steps(reducing convergence steps from 80000 to 1000 on CIFAR-10), and higher-quality DDPM also helps achieve better purification performance. Additionally, Joint-conditional Diffusion Purification achieves the best defensive results. This is because joint-condition control can bring the purified image closer to the original image, while images purified by the unconditional diffusion model tend to gradually deviate from the original clean image as the purification step increases. We show a visual example in \cref{fig:abl}.

\section{Conclusions}
To systematically investigate the vulnerability of unexploitable data, we formally define a new threat called \textit{learnable examples}, which can turn unlearnable examples into learnable. This is realized by a novel joint-condition diffusion purification process that projects the unlearnable examples onto a learnable data manifold, which is learned from other newly collected (unprotected) data. Notably, LE is independent of training scheme and consistently effective for both unauthorized supervised\&unsupervised learning. More generally, we call for future work to design UE methods that are not influenced by other unprotected data and use our approach to evaluate their performance. Because it is impractical to expect all the data in the world to be added ``unlearnable'' perturbations. In addition, such a UE solution can only slightly reduce the effectiveness of LE since LE can tolerate distribution mismatch to a great extent.    

\begin{acks}
This project has received funding from the National Natural Science Foundation of China under grant 61932009, Fundamental Research Funds for the Central Universities (No.JZ2023HGTA0202, No.JZ2023HGQA0101).
\end{acks}

\bibliographystyle{ACM-Reference-Format}
\bibliography{ref}
\end{document}